\documentclass[twoside]{article}

\usepackage[preprint]{aistats2026}
\usepackage{amssymb,amsfonts}
\usepackage{tikz}
\usetikzlibrary{arrows.meta,positioning}

\usepackage{algorithm}
\usepackage{algpseudocode}

\usepackage{booktabs,multirow,graphicx,makecell,array}
\newcommand{\rotds}[1]{\rotatebox[origin=c]{45}{\textbf{#1}}}
\usepackage{multirow}
\usepackage{textcomp}
\usepackage{xcolor}
\usepackage{flushend}
\usepackage{enumitem}
\usepackage[round]{natbib}

\begin{document}

%

%


\aistatstitle{Adaptive Heterogeneous Mixtures of Normalising Flows for Robust Variational Inference}

\aistatsauthor{ Benjamin Wiriyapong \And Oktay Karakuş \And Kirill Sidorov}

\aistatsaddress{ \textit{School of Computer Science and Informatics},
\textit{Cardiff University}, CF24 4AG, UK. \\
\{wiriyapongb, karakuso, sidorovk\}@cardiff.ac.uk} 

\begin{abstract}
    Normalising-flow variational inference (VI) can approximate complex posteriors, yet single-flow models often behave inconsistently across qualitatively different distributions. We propose \emph{Adaptive Mixture Flow Variational Inference} (AMF\mbox{-}VI), a heterogeneous mixture of complementary flows (\textsc{MAF}, \textsc{RealNVP}, \textsc{RBIG}) trained in two stages: (i) sequential expert training of individual flows, and (ii) adaptive \emph{global} weight estimation via likelihood-driven updates, without per-sample gating or architectural changes. Evaluated on six canonical posterior families of \textit{banana, X-shape, two-moons, rings, a bimodal,} and \textit{a five-mode mixture}, AMF\mbox{-}VI achieves consistently lower negative log-likelihood than each single-flow baseline and delivers stable gains in transport metrics (Wasserstein-2) and maximum mean discrepancy (MDD), indicating improved robustness across shapes and modalities. The procedure is \emph{efficient and architecture-agnostic}, incurring minimal overhead relative to standard flow training, and demonstrates that adaptive mixtures of diverse flows provide a reliable route to robust VI across diverse posterior families whilst preserving each expert’s inductive bias.
\end{abstract}

\section{Introduction}

Variational inference (VI) is widely used for approximate Bayesian computation, yet complex \emph{multimodal} posteriors remain challenging. Classical VI with simple families (e.g., Gaussians) lacks the capacity to represent multiple modes, leading to underestimated uncertainty and poor fits \citep{bishop2006pattern}. Normalising flows increase expressiveness \citep{kobyzev2020normalizing,papamakarios2021normalizing}, but individual architectures often specialise to particular structures and can suffer mode collapse during optimisation \citep{che2016mode}, limiting reliability in applications such as generative modelling, Bayesian neural networks, and hierarchical posteriors.

Multimodality is a \emph{practical} requirement in many domains: hierarchical Bayesian models can yield distinct epidemiological regimes \citep{wakefield2007disease}, deep generative models often induce discrete semantic clusters \citep{kingma2013auto}, and financial returns switch across market regimes \citep{ang2002regime,hamilton1989new}. While normalising flows have shown strong capacity to model such structure \citep{kobyzev2020normalizing,papamakarios2021normalizing}, single–flow architectures face a bias–variance tension when asked to cover diverse modes with one parametric family. This motivates mixture–based flows that can \emph{adaptively} allocate capacity across modes while remaining computationally tractable for variational inference \citep{blei2017variational}.

Current mixture–enhanced flows typically use \emph{global, data–independent} mixing weights $\{\pi_k\}$ learned once and fixed at test time \citep{PiresFigueiredo2020VMoNF,NgZammitMangion2024SphereMixFlow,Izmailov2019FlowGMM,Kobyzev2021NFReview,Bishop2006PRML,Dempster1977EM}. Such global weighting struggles to reallocate mass to local complexities (secondary modes, tails), a problem exacerbated under the \emph{reverse} divergence $\mathrm{KL}(q\|p)$, which is mode–seeking and can under–disperse \citep{Blei2017VIReview}. Although tail–aware VI objectives \citep{Wang2018TailAdaptive} and base–distribution tweaks \citep{Stimper2022ResampledBase} help, a gap remains between the \emph{adaptive} nature of targets and the \emph{global} structure of many mixture–of–flows designs, where tractable \emph{dynamic} allocation across the full support is still challenging \citep{Kobyzev2021NFReview}. Related experience with mixture density networks reports instability and mode under–representation without extra mechanisms \citep{Bishop2006PRML,Makansi2019MDNLimitations}, underscoring the need for mixtures that \emph{adapt allocation} rather than fixing it globally.

Despite substantial progress in normalising\mbox{-}flow architectures—e.g., RealNVP \citep{dinh2016realnvp}, Masked Autoregressive Flow (MAF) \citep{papamakarios2017masked}, and Rotation\mbox{-}based Iterative Gaussianisation (RBIG) \citep{laparra2011iterative}—there remains a need for a principled, tractable way to \emph{combine} heterogeneous flows for reliable multimodal posterior inference across diverse geometries. Individual flows offer complementary strengths (RealNVP’s efficient coupling layers for high\mbox{-}dimensional transforms, MAF’s strong density estimation, RBIG’s robust iterative Gaussianisation), suggesting that adaptive component selection could improve approximation quality \citep{rezende2015variational,chen2018neural}. However, existing mixture approaches in VI often use fixed global weights or heuristic adaptation, which under\mbox{-}exploit the interplay between component specialisation and target structure \citep{tomczak2018vae,berg2018sylvester}. Moreover, current work on mixture normalising flows has not fully addressed learning data\mbox{-}driven \emph{optimal} combinations, nor explored sequential training strategies that leverage meta\mbox{-}learning to improve coordination and convergence \citep{finn2017model,nichol2018first}.

The absence of sophisticated frameworks for adaptive sampling from learned mixture components represents a critical gap in the current literature, particularly given the potential for dynamically adjusting component importance based on local distribution complexity and sampling requirements \citep{loaiza2019continuous, behrmann2019invertible}. Existing approaches to mixture model adaptation in related domains, such as mixture of expert architectures \citep{shazeer2017outrageously} and ensemble methods \citep{lakshminarayanan2017simple}, have not been systematically adapted to the unique challenges posed by normalising flow mixtures, where maintaining invertability and tractable density computational challenges. Additionally, the lack of comprehensive analysis regarding how different flow architectures specialise within mixture context, including their respective abilities to capture specific modes, handle varying local curvatures, and maintain stable training dynamics, represents a significant obstacle to developing principled mixture design strategies which could substantially advance the state of variational inference for complex, multimodal distributions.    

\paragraph{Motivation and aims.}
This work develops an \emph{adaptive mixture} framework for normalising flows that tackles a core limitation of single-flow architectures: brittle behaviour across qualitatively different posterior geometries. Our objective is to combine heterogeneous inductive biases through \emph{data-driven global weighting} while keeping the training pipeline lightweight and reproducible. Concretely, we investigate (i) \emph{sequential} training protocols in which individual flows specialise on complementary aspects of the target before combination, connecting to multi–stage learning and curriculum ideas \citep{bengio2009curriculum} and ensemble methodology \citep{breiman2001random}; (ii) \emph{non–parametric} weight adaptation based on likelihood–driven moving–average updates that adjust component importance without per–sample gating, informed by mixture–of–experts principles \citep{jacobs1991adaptive} and adaptive learning frameworks \citep{sutton1998reinforcement}; and (iii) rigorous evaluation beyond log–likelihood, incorporating distributional distance and transport–theoretic metrics \citep{villani2008optimal}. The goal is a practical, architecture–agnostic recipe that yields robust posterior approximation across diverse families while retaining the simplicity of standard flow training.

\paragraph{Contributions.}
We introduce \emph{Adaptive Mixture Flow Variational Inference} (AMF\mbox{-}VI), a heterogeneous mixture of normalising flows (MAF, RealNVP, RBIG) trained in two stages: \textit{Stage~1} trains experts \emph{sequentially} and independently; \textit{Stage~2} estimates \emph{global} mixture weights post hoc via likelihood–driven moving–average updates (softmax–normalised), with a light smoothing/floor to avoid degenerate collapse. No per–sample gating or end–to–end joint training is required. We evaluate on six canonical posterior families, \textit{banana, X–shape, two–moons, rings, a bimodal,} and \textit{a five–mode mixture} and report five complementary metrics: negative log–likelihood (NLL), Kullback–Leibler (KL) divergence, Wasserstein--2 ($W_2$) distance, and maximum mean discrepancy (MDD) in unbiased and biased forms. Empirically, AMF\mbox{-}VI achieves consistently lower NLL than each single–flow baseline across these families and exhibits stable improvements in $W_2$/MDD, indicating enhanced robustness to shape and modality variation under a minimal training and implementation footprint.

The rest of the paper is organised as follows: Section \ref{sec:methods} formulates \emph{AMF-VI}, where Section \ref{sec:data_eval} presents dataset details and evaluation metrics. Section \ref{sec:results} presents quantitative and qualitative results with ablations. Section \ref{sec:conc} discusses limitations and computational trade-offs and concludes the paper.

\section{Methodology}\label{sec:methods}
This section presents AMF\mbox{-}VI, a two–stage, heterogeneous mixture of normalising flows for posterior approximation. In \textbf{Stage~1}, architecturally diverse experts (e.g., \textsc{MAF}, \textsc{RealNVP}, \textsc{RBIG}) are trained \emph{independently}, where in \textbf{Stage~2}, with expert parameters fixed, we estimate \emph{global} mixture weights by a likelihood–driven moving–average update, yielding a tractable, data–agnostic gate that reallocates mass across experts without per–sample gating. 
The remainder of this section formalises the mixture, details weight learning, motivates the moving–average scheme, and summarises the expert classes.

\subsection{Mixture Model Formulation}

\subsubsection{Mixture Posterior Approximation}
The central construct in AMF-VI is the mixture posterior approximation, defined as:
\begin{align}
    q_\phi(z) = \sum_{k=1}^K \pi_k q_k(z|\phi_k)
\end{align}  
where $z \in \mathbb{R}^d$ represents the latent variables, $K$ denotes the number of mixture components, $\pi_k$ are the mixing weights satisfying $\sum_{k=1}^K \pi_k = 1$ and $\pi_k \geq 0$ for all $k$, and $q_k(z|\phi_k)$ represent the $k$-th component density parameterised by $\phi_k$ \citep{rezende2015variational, kingma2013auto}. 

Each component $q_k(z|\phi_k)$ is typically implemented as a normalising flow, enabling the transformation of a simple base distribution (usually standard Gaussian) into a more complex, expressive distribution through a sequence of invertible mapping \citep{papamakarios2021normalizing}. This formulation allows the mixture to capture multi-modal posterior distributions that single-component variational approximations often fail to represent adequately.

\subsubsection{Role of Mixing Weights and Component parameters}
The mixing weights $\pi_k$ serve a dual purpose in the AMF-VI framework. First, they determine the relative importance of each component in the mixture, effectively controlling the probability mass allocated to different regions of the latent space \citep{bishop2006pattern}. Second, in the adaptive setting, these weights can be dynamically adjusted during training to focus computational resources on the most relevant components, leading to improved efficiency and convergence properties \citep{tomczak2018vae}.

\subsection{Learning the Mixing Weights}
Mixture models typically learn component weights alongside component parameters through joint optimisation. However, when combining flows with heterogeneous architectures, such as parametric autoregressive models and non-parametric iterative transformations, joint training requires careful balancing of learning rates and convergence criteria across fundamentally different optimisation procedures. Our sequential AMF-VI framework addresses this practical challenge through a two-stage training methodology: first, each flow expert specialises independently; second, mixture weights adapt through a moving average of component performances. This decoupling achieves robust specialisation without the complexity of coordinating heterogeneous optimisation procedures.     

\subsubsection{Sequential Training Architecture}
Our Sequential AMF-VI framework employs a two-stage training methodology that differs from traditional responsibility-weighted approaches. Our method uses sequential training to achieve flow specialisation through temporal separation and performance-based weight adaptation.

The mixture weights $\pi_k$ are constrained to the probability simplex, satisfying $\sum_{k=1}^K \pi_k = 1$ and $\pi_k \geq 0$ for all $k$. Rather than parameterising these weights through unconstrained logits and applying softmax transformations, our approach directly updates the weights in the probability simplex through performance-weighted moving averages. This direct parameterisation maintains numerical stability and ensures automatic normalisation throughout the training process.

The key departure from traditional mixture training lies in the temporal separation of flow parameter optimisation and weight training. This approach prevents the instability that can arise from simultaneous optimisation of both flow parameters and mixture weights. particularly when dealing with heterogeneous flow architectures with vastly different convergence characteristics \citep{bishop2006pattern}

\subsubsection{Two-Stage Training Dynamics}
Our training procedure consists of two distinct phases that enable effective specialisation without the computational overhead of full responsibility-weighted training:

\emph{Stage 1: Independent Flow Specialisation}: Each flow expert $q_k$ is trained independently to maximise its individual log-likelihood:
\begin{align}
    \theta_k^* = \arg\max_{\theta_k} \mathbb{E}_{z \sim q_k} 
    [\log q_k(z | \theta_k)]
\end{align}

The implementation accommodates both parametric flows (MAF, RealNVP) through standard gradient-based optimisation and non-parametric flow (RBIG) through specialised fitting procedures. This stage establishes diverse initial representations by allowing each flow to develop its optimal parameters independently, ensuring that different architectural biases lead to complementary specialisations.

\emph{Stage 2: Performance-Weighted Moving Average Adaptation}: With flow parameters fixed at $\theta_k^*$, the mixture weights are updated directly in the probability simplex using an exponential moving average of normalised flow performances:
\begin{align}
    \pi_k^{(t+1)} = \beta \pi_k^{(t)} + (1-\beta) \cdot 
    \frac{\exp(\ell_k^{(t)})}{\sum_{j=1}^K \exp(\ell_j^{(t)})}
\end{align}
where $\ell_k^{(t)} = \mathbb{E}_{z \sim p_{\text{fresh}}} [\log q_k(z | \theta_k^*)]$ represents the average log-likelihood of the $k$-th flow on freshly sampled data, and $\beta = 0.9$ is the momentum parameter. The softmax normalisation of log-likelihoods $\exp(\ell_k^{(t)}) / \sum_j \exp(\ell_j^{(t)})$ produces performance-based target weights that favour flows with higher likelihoods, whilst the moving average provides temporal smoothing to prevent rapid weight oscillations. This update rule maintains $\sum_k \pi_k^{(t+1)} = 1$ automatically through the convex combination of normalised distributions.

\subsection{Advantages of Moving Average Weight Adaptation}
The moving average weight adaptation mechanism provides several critical advantages over traditional gradient-based optimisation approaches:
\begin{enumerate}[topsep=0pt,itemsep=-1ex,partopsep=1ex,parsep=1ex]
    \item \emph{Computational Efficiency}: The absence of gradient computation and back-propagation through mixture weights significantly reduces training time and memory requirement compared to joint optimisation schemes
    \item \emph{Numerical Stability}: The exponential smoothing formulation eliminates gradient explorations or vanishing gradients in weight learning, preventing rapid oscillations that can destabilise mixture training, particularly when combining flows with disparate numerical characteristics
    \item \emph{Interpretability}: Mixture weights directly reflect individual flow performance, enabling transparent understanding of component contributions to the overall approximation
    \item \emph{Fresh Data Evaluation}: Using a newly generated sampler for weight updates prevents overfitting to specific training batches and provides unbiased performance estimates of each flow's generation capability
    \item \emph{Trade-off with Convergence Guarantees}: Unlike joint EM optimisation, this approach does not guarantee convergence to the same local optima. However, the sequential nature avoids the practical challenges of balancing learning rates between heterogeneous flow architectures, which often require extensive hyper-parameter tuning in joint training \citep{mclachlan2008algorithm}.
\end{enumerate}
    


These advantages make the moving average approach particularly suitable for heterogeneous mixture models where flows have disparate convergence behaviours and architectural properties. The effectiveness of this framework depends critically on the diversity of the component flows, which we detail in the following section.

\subsection{Heterogeneous Flow Experts}
Our framework combines three architecturally distinct normalising flows, each bringing complementary inductive biases:

\textbf{Masked Autoregressive Flow (MAF)} \citep{papamakarios2017masked} 
models dependencies through autoregressive transformations:
\begin{align}
z_i = \mu_i(\mathbf{z}_{<i}; \theta) + \sigma_i(\mathbf{z}_{<i}; \theta) \odot \epsilon_i
\end{align}
where $\mathbf{z}_{<i} = (z_1, \ldots, z_{i-1})$ denotes previous dimensions 
and $\epsilon_i \sim \mathcal{N}(0,1)$.

\textbf{RealNVP} \citep{dinh2016realnvp} employs affine coupling layers:
\begin{align}
\mathbf{z}_{1:d} &= \mathbf{z}_{1:d}\\
\mathbf{z}_{d+1:D} &= \mathbf{z}_{d+1:D} \odot \exp(s(\mathbf{z}_{1:d}; \theta)) 
+ t(\mathbf{z}_{1:d}; \theta)
\end{align}
where $s(\cdot)$ and $t(\cdot)$ are scale and translation networks.

\textbf{RBIG} \citep{laparra2011iterative} alternates rotation and 
marginal Gaussianisation:
\begin{align}
\mathbf{z}^{(l+1)} = R^{(l)} \circ G^{(l)}(\mathbf{z}^{(l)})
\end{align}
where $G^{(l)}$ applies marginal Gaussianisation and $R^{(l)}$ is an 
orthogonal rotation.

The three flows provide complementary modelling capabilities through their distinct architectural constraints. MAF's autoregressive structure enforces explicit sequential dependencies between dimensions, RealNVP's coupling layers enable efficient parallel computation through dimension partitioning, and RBIG's non-parametric approach adapts to arbitrary marginal distributions without parametric assumptions. 

The sequential training framework allows each flow to converge independently before weight adaptation begins, avoiding the practical challenges of balancing learning rates across architectures with different optimisation characteristics. The moving average weight updates, then allocate mixture responsibility based on empirical performance, creating an ensemble that 
leverages the strengths of different architectural priors. 



\subsection{Adaptive Mixture Flow Variational Inference (AMF-VI)}
\label{sec:amfvi}
AMF\mbox{-}VI is a two–stage heterogeneous mixture of normalising flows. In \textbf{Stage~1}, experts (\textsc{MAF}, \textsc{RealNVP}, \textsc{RBIG}) are trained \emph{independently} to specialise; in \textbf{Stage~2}, with expert parameters frozen, \emph{global} mixture weights $\pi$ are updated by a likelihood–driven moving average on fresh data. 
Algorithm~\ref{alg:amfvi} summarise the procedure; the following paragraphs provide details.

\begin{algorithm}[t]
\caption{AMF\mbox{-}VI: two-stage training with global weights}
\label{alg:amfvi}
\begin{algorithmic}[1]
\Require Experts $\{q_k(z\mid\theta_k)\}_{k=1}^K$, generator $p_{\text{data}}$, momentum $\beta\!\in[0,1)$, epochs $T$
\Statex \textbf{Stage~1: expert specialisation (independent training)}
\For{$k \gets 1 \,\ldots\, K$}
  \While{not converged}
    \State Sample batch $z \sim p_{\text{data}}$
    \State $\theta_k \gets \theta_k + \eta \,\nabla_{\theta_k}\,\log q_k(z\mid\theta_k)$ \Comment{RBIG uses its own fitting}
  \EndWhile
  \State Freeze $\theta_k^\ast \gets \theta_k$
\EndFor
\State Initialise $\pi^{(0)} \gets \mathrm{Uniform}(K)$
\Statex \textbf{Stage~2: global weight adaptation (moving average on simplex)}
\For{$t \gets 0 \,\ldots\, T{-}1$}
  \State Sample fresh batch $z \sim p_{\text{fresh}}$
  \For{$k \gets 1 \,\ldots\, K$}
     \State $\ell_k^{(t)} \gets \frac{1}{B}\sum_{z}\log q_k(z\mid\theta_k^\ast)$ \Comment{average log-likelihood}
  \EndFor
  \State $w \gets \mathrm{softmax}(\ell^{(t)})$
  \State $\pi^{(t+1)} \gets \beta\,\pi^{(t)} + (1{-}\beta)\,w$ \Comment{EMA; stays on simplex}
\EndFor
\State \Return $q(z)=\sum_k \pi_k^{(T)} q_k(z\mid\theta_k^\ast)$ \hfill \Comment{$N_{\mathrm{eff}}=\exp(H(\pi^{(T)}))$}
\end{algorithmic}
\end{algorithm}


The core innovation lies in the two-stage training dynamics, wherein each flow expert-MAF, RealNVP, and RBIG-first undergoes independent optimisation to develop distinct representational capacities. MAF excels at capturing autoregressive dependencies through its sequential structure, RealNVP provides computational efficiency via coupling layer transformations, whilst RBIG offers non-parametric robustness for heavy-tailed and outlier-prone distributions. Following this specialisation phase, the framework employs a moving average weight adaptation mechanism that uses fresh data evaluation to prevent overfitting and ensure generalisation-based weight allocation.

The temporal separation of flow parameter optimisation and weight learning prevents the instability commonly associated with joint optimisation of heterogeneous architectures, particularly when combining parametric and non-parametric approaches with fundamentally different convergence characteristics. This sequential approach, combined with performance-based weight allocation through a softmax-parameterised gating function, enables the emergence of natural specialisation patterns where each component assumes responsibility for modelling the distributional aspect where it demonstrates superior performance.

The resulting framework effectively implements a form of Bayesian model averaging, where the mixture automatically selects the most appropriate architectural prior for different regions of the latent space. Through this adaptive combination of complementary inductive biases, autoregressive assumptions, coupling layer constraints, and iterative Gaussianization-AMF-VI, the algorithm achieves a robust posterior approximation that transcends the limitations of individual flow architectures while maintaining computational tractability and numerical stability throughout the training process.

\section{Datasets and Evaluation Protocol}
\label{sec:data_eval}
We evaluate six canonical 2D posterior families that span smooth, non-convex, and highly multimodal geometries: \emph{Banana} (warped Gaussian with curved level sets), \emph{X\mbox{-}shape} (crossing modes with anisotropic covariance), \emph{Bimodal} (two well-separated modes), \emph{Multimodal} (five\mbox{-}component mixture), \emph{Two\mbox{-}moons} (nonlinear manifold support), and \emph{Rings} (concentric annuli). Each family is instantiated by a fixed data generator; the same generator is used for all methods. We provide three example visualisations in the first column of Fig.~\ref{fig:visuals}. For each dataset, we draw independent training and evaluation samples from the generator and report metrics on held-out evaluation samples. Unless otherwise stated, we report the mean performance over repeated runs (with different random seeds) and use a single representative seed for qualitative figures.
All baselines use their standard training losses and architectures; hyper-parameters (epochs, batch size, learning rate) are kept constant across datasets to isolate shape effects.

We report the \emph{effective number of experts} $N_{\text{eff}}=\exp(H(w))$ computed from the learned global weights to quantify specialisation in Fig.~\ref{fig:weights}. We assess complementary aspects of fit using NLL, $\mathrm{KL}(p\|q)$, Wasserstein--2 ($W_2$), and MMD (unbiased/biased).
Table~\ref{tab:metric_defs} summarises definitions and motivation behind the choice of these metrics.

\begin{table}[ht]
  \centering
  \caption{Metrics, expressions, and rationale. $^*$ ``u'' = unbiased U\mbox{-}statistic; ``b'' = biased V\mbox{-}statistic (lower variance).}
  \label{tab:metric_defs}
  \setlength{\tabcolsep}{3pt}
  \scriptsize
  \begin{tabular}{@{}p{3.1cm}l p{5.2cm}@{}}
    \toprule
    \textbf{Metric} & \textbf{Expression} & \textbf{Why use here} \\
    \midrule
    NLL & $\displaystyle -\,\mathbb{E}_{z\sim p}\big[\log q(z)\big] \;=\; H(p)+\mathrm{KL}(p\|q)$ 
        & Proxy for the VI objective; penalises mode dropping (via $\mathrm{KL}(p\|q)$). \\
    \addlinespace[2pt]
    $\mathrm{KL}(p\|q)$ & $\displaystyle \mathbb{E}_{z\sim p}\!\left[\log\frac{p(z)}{q(z)}\right]$
        & Mass\mbox{-}covering divergence aligned with ELBO optimisation. \\
    \addlinespace[2pt]
    $W_2$ & $\displaystyle \Big(\inf_{\gamma\in\Pi(p,q)} \mathbb{E}_{(x,y)\sim\gamma}\|x-y\|_2^2\Big)^{\!1/2}$
        & Transport/geometry aware; captures shape and displacement. \\
    \addlinespace[2pt]
    MMD (u/b)$^*$ & $\displaystyle \mathrm{MMD}^2_k(p,q)=\mathbb{E}_{p,p}k+\mathbb{E}_{q,q}k-2\,\mathbb{E}_{p,q}k$
        & Nonparametric two\mbox{-}sample discrepancy complementary to NLL/KL/OT. \\
    \bottomrule
  \end{tabular}
\end{table}

\section{Experimental Analysis}\label{sec:results}
In this section, we assess the robustness and consistency of \emph{AMF-VI} by comparing it against single-flow baselines, across six canonical posterior families which span unimodal, curved, and multimodal geometries to stress-test shape robustness. Performance is reported under five complementary metrics, and complete results are summarised in \ref{tab:results}.


We organise the quantitative analysis into three parts: (i) \emph{overall likelihood}, assessed by NLL; (ii) \emph{transport behaviour}, via Wasserstein--2; and (iii) \emph{divergences/discrepancies}, using $\mathrm{KL}(p\|q)$ and MMD, to expose complementary facets of performance across datasets.

\paragraph{Overall likelihood.}
Across six qualitatively different posterior families, \textbf{AMF\mbox{-}VI} attains the lowest negative log\mbox{-}likelihood (NLL) on \emph{every} dataset in Table~\ref{tab:results}: 3.463 (banana), 3.295 (X\mbox{-}shaped), 3.154 (bimodal), 3.429 (five\mbox{-}mode), 1.188 (two\mbox{-}moons), and 2.585 (rings). In contrast, single\mbox{-}flow baselines fluctuate with geometry: \textsc{MAF} diverges on \emph{banana}, \emph{bimodal} and five\mbox{-}mode (NLL $=\infty$), \textsc{RealNVP} lags on \emph{two\mbox{-}moons} and \emph{rings} (2.343 and 3.349), and \textsc{RBIG} trails on \emph{banana} (4.131). Because NLL directly reflects the variational objective, these results support the claim of \emph{shape\mbox{-}robust} inference with a single configuration, while highlighting the brittleness of individual flow families across heterogeneous posteriors~\citep{dinh2016realnvp,papamakarios2017masked}.

\paragraph{Transport behaviour.}
On Wasserstein\mbox{--}2 distance, \textbf{AMF\mbox{-}VI} is best on \emph{X\mbox{-}shaped} (0.254 vs.\ 0.286/0.364/0.302), the \emph{five\mbox{-}mode} mixture (0.218 vs. \ 0.231/0.226/0.267) and \emph{two\mbox{-}moons} (0.107 vs.\ 0.116/0.114/0.122), and \emph{near\mbox{-}best} on \emph{bimodal} (0.174, second to \textsc{RBIG} at 0.169). Specialised baselines occasionally win on specific geometries, \textsc{MAF} on \emph{banana} and \emph{five\mbox{-}mode}; \textsc{RBIG} on \emph{rings} and \emph{bimodal}, but \emph{AMF\mbox{-}VI} avoids the sharp regressions seen elsewhere and typically ranks first or second, with \emph{banana} being the main exception (third). This aligns with known inductive biases: coupling flows and Gaussianization can excel in niches yet underperform on other shapes~\citep{dinh2016realnvp,laparra2011iterative,villani2008optimal}.

\paragraph{Divergences and discrepancies.}
KL and MMD trends corroborate the likelihood picture. \textbf{AMF\mbox{-}VI} achieves the lowest $\mathrm{KL}(p\|q)$ on four out of six data sets with \emph{X\mbox{-}shaped} (0.037), \emph{bimodal} (0.024), and \emph{five\mbox{-}mode} (0.029), and ties \textsc{RealNVP} on \emph{rings} (0.042); it is close but not best on \emph{two\mbox{-}moons} (0.018 vs.\ \textsc{MAF}'s 0.014) and \emph{banana} (0.047 vs.\ \textsc{MAF}'s 0.035). For MMD, \emph{AMF\mbox{-}VI} is best on \emph{bimodal} (u/b: 0.004/0.026), \emph{five\mbox{-}mode} (0.015/0.032), and \emph{two\mbox{-}moons} (0.003/0.017), and second on \emph{X\mbox{-}shaped} (0.031/0.042), \emph{five\mbox{-}mode} (0.023/0.037) and \emph{rings} (0.037/0.046); \emph{banana} remains a favourable case for \textsc{MAF}. Taken together with uniformly superior NLL, these findings indicate that a heterogeneous mixture adapts component usage to target geometry, delivering reliable performance across shape regimes while single\mbox{-}flow baselines oscillate between strong and weak modes (or fail outright)~\citep{papamakarios2017masked,laparra2011iterative}.

\begin{table}[ht]
  \centering
  \caption{Quantitative results on six posterior families. \textbf{Bold} and \underline{underlined} values denote best and second best, respectively.}
  \label{tab:results}
  \setlength\tabcolsep{3.5pt}
  \scriptsize
  \begin{tabular}{>{\centering\arraybackslash}m{12mm} l c c c c}
    \toprule
    \textbf{Dataset} & \textbf{Metric} & \emph{AMF-VI} & \textsc{RealNVP} & \textsc{MAF} & \textsc{RBIG} \\
    \midrule
    \multirow{5}{*}{\rotds{Banana}} 
      & NLL        & \textbf{3.463} & \underline{4.026} & $\infty$ & 4.131 \\
      & KL         & 0.047          & \underline{0.044} & \textbf{0.035} & 0.066 \\
      & W2         & 0.267          & \underline{0.251} & \textbf{0.222} & 0.308 \\
      & MDD-u      & 0.031          & \underline{0.026} & \textbf{0.018} & 0.039 \\
      & MDD-b      & 0.042          & \underline{0.038} & \textbf{0.034} & 0.048 \\
    \midrule
    \multirow{5}{*}{\rotds{X-shaped}} 
      & NLL        & \textbf{3.295} & \underline{4.048} & 4.795 & 4.049 \\
      & KL         & \textbf{0.037} & \underline{0.040} & 0.044 & 0.077 \\
      & W2         & \textbf{0.254} & \underline{0.286} & 0.364 & 0.302 \\
      & MDD-u      & \underline{0.031} & \textbf{0.024} & 0.036 & 0.049 \\
      & MDD-b      & \underline{0.042} & \textbf{0.037} & 0.046 & 0.056 \\
    \midrule
    \multirow{5}{*}{\rotds{Bimodal}} 
      & NLL        & \textbf{3.154} & 3.839 & $\infty$ & \underline{3.589} \\
      & KL         & \textbf{0.024} & 0.036 & 0.035 & \underline{0.025} \\
      & W2         & \underline{0.174} & 0.195 & 0.199 & \textbf{0.169} \\
      & MDD-u      & \textbf{0.004} & 0.010 & 0.014 & \underline{0.009} \\
      & MDD-b      & \textbf{0.026} & \underline{0.028} & 0.031 & 0.029 \\
    \midrule
    \multirow{5}{*}{\rotds{Multimodal}} 
          & NLL   & \textbf{3.429} & \underline{3.852} & $\infty$   & 4.096 \\
          & KL    & \textbf{0.029} & \underline{0.037} & 0.038 & 0.050 \\
          & Wasserstein & \textbf{0.218} & 0.231 & \underline{0.226} & 0.267 \\
          & MDD-un & \textbf{0.023} & \underline{0.025} & 0.029 & 0.044 \\
          & MDD-biased & \textbf{0.037} & \underline{0.038} & 0.040 & 0.052 \\
    \midrule
    \multirow{5}{*}{\rotds{Two-moons}} 
          & NLL   & \textbf{1.188} & 2.343 & \underline{1.661} & 1.741 \\
          & KL    & 0.018 & 0.020 & \textbf{0.014} & 0.049 \\
          & W2    & \textbf{0.104} & 0.116 & \underline{0.110} & 0.124 \\
          & MDD-u & \textbf{0.000} & 0.011 & \underline{0.005} & 0.011 \\
          & MDD-b & \textbf{0.017} & 0.027 & \underline{0.021} & 0.026 \\
    \midrule
    \multirow{5}{*}{\rotds{Rings}} 
      & NLL        & \textbf{2.585} & 3.349 & 3.549 & \underline{3.263} \\
      & KL         & \textbf{0.042} & \textbf{0.042} & 0.088 & \underline{0.063} \\
      & W2         & \underline{0.354} & 0.427 & 0.519 & \textbf{0.295} \\
      & MDD-u      & \underline{0.037} & 0.057 & 0.085 & \textbf{0.008} \\
      & MDD-b      & \underline{0.046} & 0.064 & 0.090 & \textbf{0.030} \\
    \bottomrule
  \end{tabular}
\end{table}

\begin{figure}[ht]
    \centering
    \includegraphics[width=0.8\linewidth]{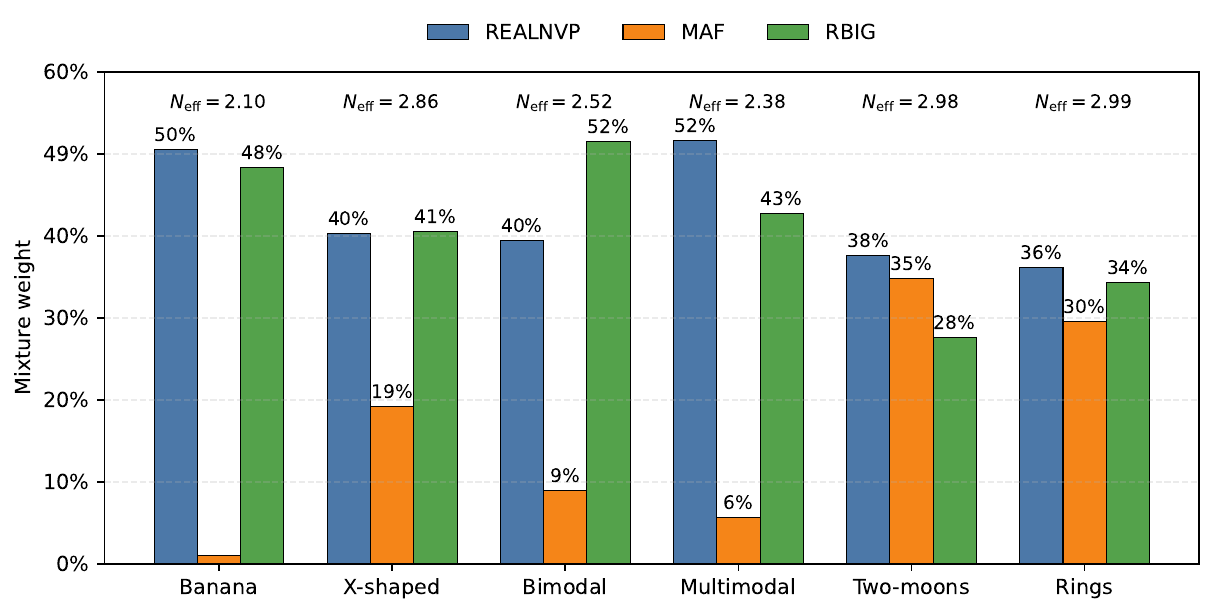}
\caption{Learned mixture weights per dataset for \emph{AMF-VI} across the three flow components (\textsc{RealNVP}, \textsc{MAF}, \textsc{RBIG}).}
\label{fig:weights}
\end{figure}

Figure~\ref{fig:weights} visualises the learned mixture weights for \textsc{RealNVP}, \textsc{MAF}, and \textsc{RBIG} across the six posterior families, showing adaptation to geometry rather than collapse to a single expert. For \emph{Banana}, the gate assigns almost zero weight to \textsc{MAF} (0.010) and splits mass between \textsc{RealNVP} (0.505) and \textsc{RBIG} (0.484), consistent with \textsc{MAF}'s divergence on this dataset and the stronger baselines among the other two. On \emph{X\mbox{-}shaped}, \textsc{RealNVP} and \textsc{RBIG} receive comparable weight (0.403/0.406), aligning with the strong transport/discrepancy performance of these families, while \emph{AMF\mbox{-}VI} still achieves the best NLL/KL in Table~\ref{tab:results}. On the \emph{Bimodal} target, \textsc{RBIG} is emphasised (0.515), matching its favourable $W_2$ score (with \emph{AMF\mbox{-}VI} second), whereas \emph{AMF\mbox{-}VI} remains best on NLL/MMD. For the \emph{Multimodal} (five\mbox{-}mode) case, the gate prioritises \textsc{RealNVP} (0.517) and \textsc{RBIG} (0.427) and downweights \textsc{MAF} (0.056), reflecting \textsc{MAF}'s unstable likelihood there despite its narrow advantage on $W_2$; the mixture consequently attains the best NLL/KL and MMD. On \emph{Two\mbox{-}moons}, \textsc{RealNVP} and \textsc{MAF} receive similar weights (0.376/0.348) while \textsc{RBIG} is down-weighted (0.276), matching the stronger transport/divergence behaviour of the former pair and the best NLL/$W_2$ of \emph{AMF\mbox{-}VI}. Finally, for \emph{Rings} the allocation is broadly balanced (0.362/0.295/0.343), consistent with the absence of a single dominant baseline across metrics (with \textsc{RBIG} leading $W_2$/MMD and \emph{AMF\mbox{-}VI} leading NLL). Overall, the weights remain distributed and adjust to target geometry, indicating that the mixture exploits complementary inductive biases without component collapse, which helps explain the stable, across\mbox{-}family NLL gains in Table~\ref{tab:results}.

\begin{figure*}[t!]
    \centering
    \includegraphics[width=\linewidth]{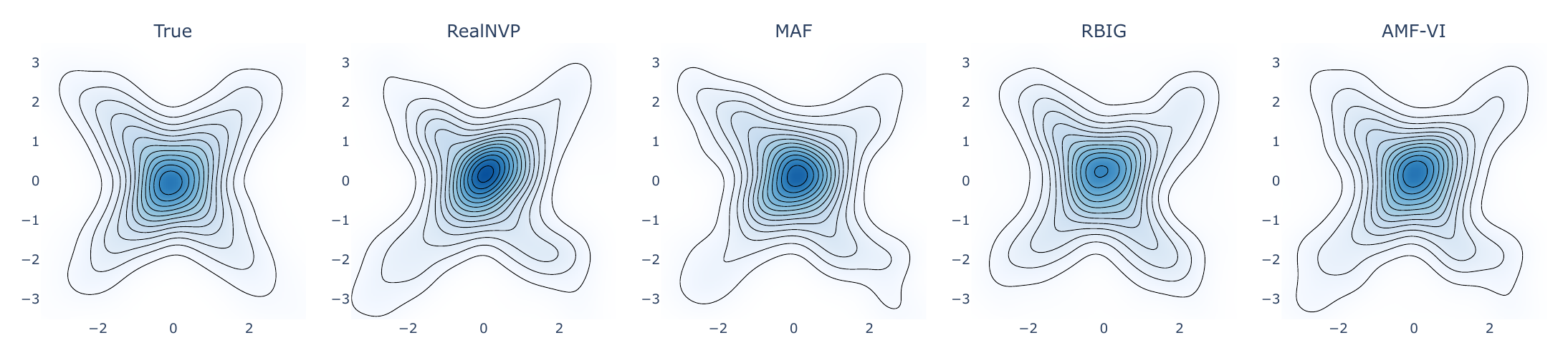}\\
    \includegraphics[width=\linewidth]{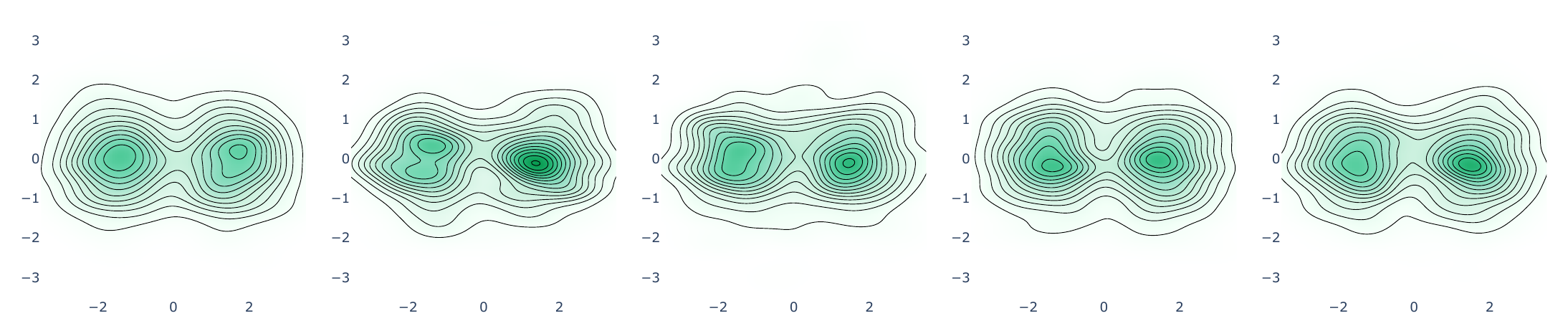}\\
    \includegraphics[width=\linewidth]{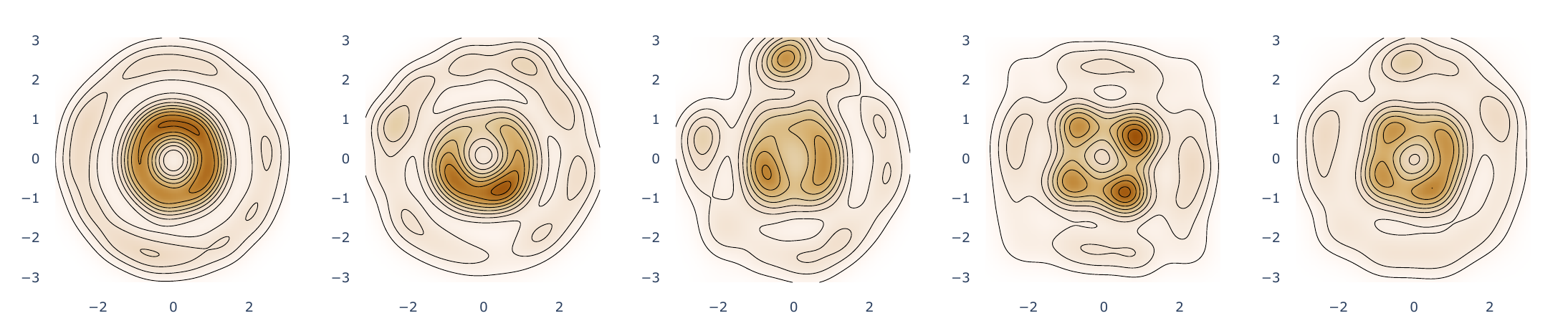}
\caption{Qualitative comparison on three posterior families, (TOP) \emph{X-shaped}, (MIDDLE) \emph{Bimodal}, and (BOTTOM) \emph{Rings}. Each subfigure shows (left$\to$right) true data and samples from \textsc{RealNVP}, \textsc{MAF}, \textsc{RBIG}, and \textbf{AMF\mbox{-}VI}.}
\label{fig:visuals}

\end{figure*}

To quantify how many experts are effectively used by the gate, we report $N_{\text{eff}}(w)=\exp\!\big(H(w)\big)$, where $H$ is the Shannon entropy of the \emph{global} mixture weights ($N_{\text{eff}}\!\in[1,K]$, here $K{=}3$; $1$ indicates collapse, $3$ a perfectly balanced mix). Across datasets, we obtain:
\emph{Banana} $2.10$, \emph{X\mbox{-}shaped} $2.86$, \emph{Bimodal} $2.52$, \emph{Multimodal} $2.38$, 
\emph{Two\mbox{-}moons} $2.98$, and \emph{Rings} $2.99$ (mean $\approx 2.64$, range $[2.10,\,2.99]$).
These values confirm \emph{no component collapse} and \emph{controlled specialization}: near\mbox{-}three on \emph{Two\mbox{-}moons}/\emph{Rings} indicates broad participation of all experts, whereas a lower value on \emph{Banana} reflects the two\mbox{-}expert allocation observed in Figure~\ref{fig:weights}.
Overall, $N_{\text{eff}}$ trends agree with the weight profiles and help explain the robust, across\mbox{-}family NLL gains.

Figure~\ref{fig:visuals} presents qualitative samples for three representative posterior families, (TOP) \emph{X-shaped}, (MIDDLE) \emph{Bimodal}, and (BOTTOM) \emph{Rings}, contrasting our \textit{AMF\mbox{-}VI} against single-flow baselines (\textsc{RealNVP}, \textsc{MAF}, \textsc{RBIG}). We read these visuals in conjunction with the quantitative results in Table~\ref{tab:results}. Evaluating Figure \ref{fig:visuals}-(TOP), \textit{AMF\mbox{-}VI} reconstructs the four arms with clean angular structure and limited spurious mass near the centre, while \textsc{RealNVP} captures the cross but is more diffuse at the tips; \textsc{MAF} visibly blurs the arms and \textsc{RBIG} concentrates too much mass centrally. This matches Table~\ref{tab:results}: \textit{AMF\mbox{-}VI has the best NLL} (3.295) and \textit{best} $W_2$ (0.254), whereas \textsc{RealNVP} achieves slightly lower MDD values (MDD-u 0.024, MDD-b 0.037) at the cost of worse NLL (4.048).

Figure \ref{fig:visuals}-(MIDDLE), \textit{AMF\mbox{-}VI} clearly resolves the two separated modes with minimal bridging mass; \textsc{RealNVP} produces broader lobes, \textsc{MAF} shows mode-bridging, and \textsc{RBIG} remains comparatively diffuse. Quantitatively, \textit{AMF\mbox{-}VI leads on NLL} (3.154), \textit{KL} (0.024), and \textit{both} MDDs (MDD-u 0.004, MDD-b 0.026), and is \emph{second-best} on $W_2$ (0.174), slightly behind \textsc{RBIG} (0.169). This indicates high fidelity at data locations together with strong global geometry. Lastly, Figure \ref{fig:visuals}-(BOTTOM), \textit{AMF\mbox{-}VI} reconstructs both rings with mild angular irregularities; \textsc{RealNVP} forms rings but with azimuthal drift; \textsc{MAF} exhibits spoke-like artefacts; \textsc{RBIG} appears “square-ish.” Despite that visual artifact, \textsc{RBIG} scores best on \emph{radial}-dominated metrics ($W_2$ 0.295; MDD-u 0.008; MDD-b 0.030), while \textit{AMF\mbox{-}VI achieves the best NLL} (2.585) and ties \textsc{RealNVP} on KL (0.042). This pattern reflects that transport and RBF–MMD emphasise radial alignment of the two rings, whereas NLL is more sensitive to local density at true data locations.

\section{Conclusion}\label{sec:conc}
We introduced \emph{AMF\mbox{-}VI}, a heterogeneous mixture of normalising flows trained by sequentially fitting expert flows (Stage~1) followed by \emph{post\mbox{-}hoc} likelihood\mbox{-}driven estimation of \emph{global} mixture weights (Stage~2). Across six canonical posterior families, banana, X\mbox{-}shape, two\mbox{-}moons, rings, a bimodal, and a five\mbox{-}mode mixture, \emph{AMF\mbox{-}VI} achieves the \emph{lowest} NLL on all datasets and remains competitive on transport/discrepancy metrics (Table~\ref{tab:results}). The learned weights are interpretable and non\mbox{-}collapsed (e.g., $N_{\text{eff}}\in[2.10,\,2.99]$, mean $\approx2.64$), indicating controlled specialisation rather than single\mbox{-}expert dominance. Qualitatively, the gate emphasises experts whose inductive biases align with geometry (e.g., \textsc{RealNVP}/\textsc{RBIG} on \emph{banana}; more balanced \textsc{RealNVP}/\textsc{MAF} on \emph{two\mbox{-}moons}), which helps explain stable performance even when a baseline fails (e.g., \textsc{MAF} divergence on \emph{banana}/\emph{bimodal} does not propagate to the mixture).

Compute/memory scale roughly linearly with the number of experts $K$; for narrowly scoped posteriors, a single deeper flow might be cheaper. Our Stage~2 uses \emph{global} (data\mbox{-}agnostic) weights rather than per\mbox{-}sample gating and can be sensitive to noisy likelihood estimates if weights become overly peaked (we mitigate with smoothing/floors). Experiments focus on low\mbox{-}to\mbox{-}moderate dimensional targets and canonical shapes; scaling to high\mbox{-}dimensional, constrained, or strongly multi\mbox{-}modal scientific posteriors (and amortised settings) remains open. Performance also depends on the expert set; adding spline/NSF or continuous\mbox{-}time flows could improve coverage at the cost of design complexity.

Promising directions include mixture\mbox{-}aware training (e.g., responsibility\mbox{-}weighted objectives) while retaining the two\mbox{-}stage pipeline, principled specialisation controls (e.g., information\mbox{-}theoretic regularisation or $N_{\text{eff}}$ targeting), and extending to amortised inference and higher\mbox{-}dimensional benchmarks. Applying \emph{AMF\mbox{-}VI} to domain\mbox{-}specific posteriors where reliability is critical is a natural next step.

\newpage
\bibliography{refs}
\end{document}